\documentclass[letterpaper, 10 pt, conference]{ieeeconf}  

\IEEEoverridecommandlockouts                              

\title{\LARGE \bf
QuadAgent: A Responsive Agent System for Vision-Language Guided Quadrotor Agile Flight
}

\author{Ao Zhuang, Feng Yu, Tianbao Zhang, Linzuo Zhang, Danping Zou$^{\dag}$
\thanks{$^{\dag}$ Corresponding author.}%
\thanks{\scriptsize \tt Emails: \{zhuangaoooo, yu-feng, zhangtianbao, zhanglinzuo, dpzou\}@sjtu.edu.cn}%
}

\usepackage{amssymb}
\usepackage{amsmath}
\usepackage{graphicx}
\usepackage{booktabs}
\usepackage{multirow}
\usepackage[T1]{fontenc}
\usepackage{lettrine}
\usepackage{graphicx}
\usepackage{amsmath}
\usepackage{amsfonts}
\usepackage{booktabs}
\usepackage{threeparttable}

\usepackage{multirow}
\usepackage[table]{xcolor}
\usepackage{colortbl}
\usepackage{makecell}
\usepackage{tabularx}
\usepackage[normalem]{ulem}

\makeatletter
\let\NAT@parse\undefined
\makeatother
\usepackage[hidelinks]{hyperref}

\begin{document}

\maketitle
\thispagestyle{empty}
\pagestyle{empty}

\begin{abstract}


We present QuadAgent, a training-free agent system for agile quadrotor flight guided by vision-language inputs. Unlike prior end-to-end or serial agent approaches, QuadAgent decouples high-level reasoning from low-level control using an asynchronous multi-agent architecture: Foreground Workflow Agents handle active tasks and user commands, while Background Agents perform look-ahead reasoning. The system maintains scene memory via the Impression Graph, a lightweight topological map built from sparse keyframes, and ensures safe flight with a vision-based obstacle avoidance network. Simulation results show that QuadAgent outperforms baseline methods in efficiency and responsiveness. Real-world experiments demonstrate that it can interpret complex instructions, reason about its surroundings, and navigate cluttered indoor spaces at speeds up to $5$ m/s.

\end{abstract}

\section{Introduction}
Recent advances in large language models (LLMs) and vision-language models (VLMs) have opened up new opportunities for autonomous robotics, enabling agents to follow high-level semantic commands and perform complex reasoning.  In aerial robotics, this promises intelligent and flexible flight guided by natural language instructions and visual contexts. 
To achieve this goal, one common approach is to train an end-to-end vision-language navigation (VLN) model which maps  language instructions and sensory inputs directly to control actions \cite{OpenFly, wang2025uavflowcolosseorealworldbenchmark}. While these methods demonstrate impressive capabilities in specialized tasks, they typically require massive quantities of high-fidelity, expert-level flight data and substantial computational resources for training. Beyond these resource requirements, the end-to-end architecture makes it inherently difficult to integrate explicit memory modules or structured scene priors. This limitation hinders the system's capacity for long-horizon spatial reasoning and efficient path planning in complex, large-scale environments.

\begin{figure}
  \centering
  \includegraphics[width=0.49\textwidth]{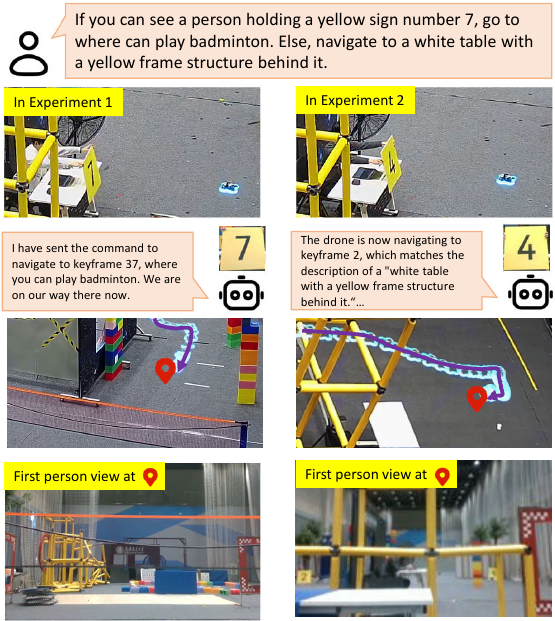}
  \caption{\textbf{Complex reasoning using our agent system.} Given the identical conditional task, the left and right panels illustrate the agent's behavior depending on observations: the agent navigates to the badminton net through random obstacles when "yellow sign number 7" is observed (left), and proceeds to the white table in front of the yellow frame when it is not (right).
}
  \label{fig:condition}
\end{figure}

Alternatively, the emerging paradigm of general-purpose agents, as demonstrated by recent breakthroughs such as Manus \cite{manus} and OpenClaw \cite{openclaw}, offers a promising approach for tackling complex tasks through autonomous tool invocation without fine-tuning. Nevertheless, applying similar architectures to UAVs presents significant challenges due to two fundamental limitations. First, conventional agent architectures, such as ReAct \cite{yao2022react}, rely on serial, blocking tool-use logic, where a function call must finish before the next action can start. Consequently, time-consuming physical actions block the reasoning cycle, making the system unresponsive to changing inputs while the UAV is in motion. Second, UAVs and foundation models have mismatched characteristics. UAVs have limited flight time, while multi-turn inference can be slow. In addition, agile 3D navigation needs to work with 2D visual reasoning. As a result, prior methods have largely focused on offline planning \cite{wu2025selp} or simulation \cite{uav-codeagent}, leaving real-world agile UAV deployment largely unsolved.

To address the conflicts between interaction fluidity, inference latency, and navigational agility, the agent system must respond to aperiodic mid-flight inputs without interrupting physical actions. It must also perform complex reasoning during flight, as illustrated in Fig. \ref{fig:condition}. This requires not only the reasoning capabilities of foundation models but also maintaining a long-term memory to store key scene information.


To meet those challenges, we propose QuadAgent, a responsive training-free agent system for agile quadrotor control.
Unlike serial stop-and-infer pipelines, where reasoning is blocked by executing flight actions \cite{pardyl2025flysearch, xiao2025uav}, QuadAgent separates high-level reasoning from low-level control, allowing them to run concurrently.

The cognitive layer consists of two types of agents: Foreground Workflow Agents and Background Agents. It uses an event-driven mechanism, treating physical skill invocations as internal state triggers rather than blocking functions. For scene memory, we introduce the Impression Graph, a lightweight topological map built from sparse keyframe images, where connectivity between keyframes encodes traversability.

At the physical layer, we incorporate a differentiable, physics-based obstacle avoidance policy \cite{zhang2025learning}, providing a robust safety net against unmapped obstacles and network delays. This design allows the agent to perform complex reasoning while maintaining agile, safe flight. 

We validate our system through both simulation and real-world experiments. The results demonstrate that our agent significantly outperforms baseline methods in navigation efficiency while maintaining robust and agile control. Our responsive asynchronous design enables non-blocking reasoning during flight. Together with lightweight scene priors and a robust collision avoidance layer, the quadrotor achieves agile navigation in cluttered indoor spaces at speeds up to 5 m/s. In summary, our contributions are threefold:
\begin{itemize}
    \item We propose QuadAgent, a training-free agent system that integrates high-level foundation models (LLM and VLM) with low-level vision-action networks to guide agile quadrotor flight using vision-language inputs. The system is flexible and can incorporate more skills and advanced models in the future.
    \item We design an asynchronous multi-agent architecture that enables responsive human-robot interactions and mitigates inference latency through background look-ahead reasoning. 
    \item We introduce the Impression Graph, a lightweight topological scene representation that enables agile 3D navigation based on 2D visual reasoning.  

\end{itemize}

\begin{figure*}
\centering
\includegraphics[width=0.95\textwidth]{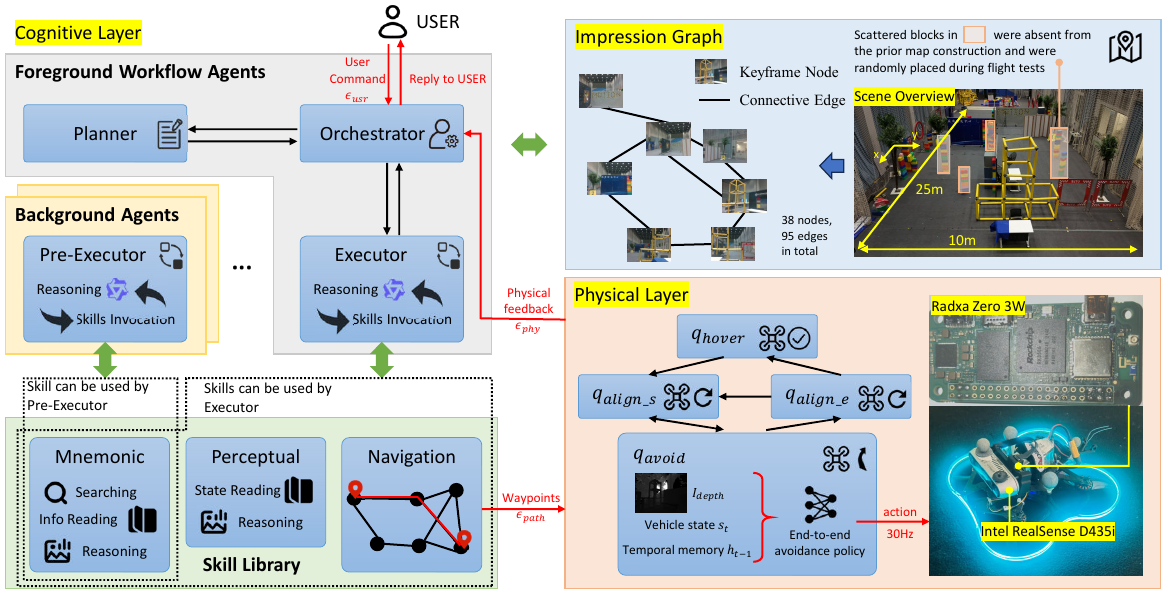}
\caption{
\textbf{System Overview.} 
In Foreground Workflow Agents, the orchestrator monitors events ($\epsilon_{usr}, \epsilon_{phy}$) in the idle state and routes tasks to the planner or executor. 
Both the executor and pre-executor autonomously call mnemonic, navigation, or perceptual skills from the skill library.
Notably, the navigation skill triggers $\epsilon_{path}$ to drive the onboard physical layer state machine for actuation, transitioning among the following states: $q_{hover}$ (hovering), $q_{align\_s}$ (in-place rotation to align with the initial path heading), $q_{avoid}$ (path tracking via collision avoidance policy), and $q_{align\_e}$ (in-place rotation to the final target orientation). The Impression Graph at the top right underpins mnemonic and navigation skills as the scene prior representation. Images in the Impression Graph and the  physical layer show the arena and UAV used in real-world experiments.
}
\label{fig:system_overview}
\end{figure*}

\section{Related Work}
Current approaches combining  LLMs or VLMs with robotics can be categorized into two primary paradigms based on whether the underlying foundation models are trained or fine-tuned.

The first strategy involves fine-tuning VLMs to generate manipulation actions \cite{pi0.5} or navigation commands \cite{navfom} in an end-to-end manner. Similar paradigms have also been extended to aerial navigation \cite{OpenFly, wang2025uavflowcolosseorealworldbenchmark}. UAV-Flow Colosseo \cite{wang2025uavflowcolosseorealworldbenchmark} establishes a large-scale  mixed dataset of real and simulated flights and fine-tunes $\pi_0$ \cite{pi0} and OpenVLA \cite{kim2024openvla} for language-guided trajectory control. As these approaches rely on imitation learning, their performance depends on the scale and diversity of expert demonstrations, which may limit generalization in agile aerial navigation. Furthermore, without explicit intermediate scene representations or persistent memory, such end-to-end architectures primarily operate in a reactive manner, making long-horizon planning  challenging.

The second strategy employs frozen LLMs or VLMs as modular components within predefined workflows \cite{hu2025see, uss-nav, li2025skyvln, airhunt}. For example, SPF \cite{hu2025see} treats action prediction as a 2D spatial grounding task. It prompts a frozen VLM to annotate 2D waypoints on the input image, which are then geometrically transformed into executable 3D displacement vectors. The primary advantage of these methods lies in exploiting the advanced reasoning capabilities of foundation models without incurring expensive training costs. However, as the overall procedure is structured around manually designed pipelines tailored to specific tasks, extending these systems to broader or more complex scenarios may require substantial redesign.

Other training-free methods develop agents capable of autonomous tool invocation \cite{pmlr-v205-ichter23a, roboos, uav-codeagent}. For instance, UAV-CodeAgents \cite{uav-codeagent} employs a multi-agent ReAct paradigm to generate flight missions in simulation. However, the standard ReAct loop within individual agents does not proceed to the next step of reasoning until the flight tool invocation finally returns the observation of the entire trajectory. This stop-and-infer pipeline with blocking tool-use logic forces the vehicle to halt for multi-turn inference before proceeding and reduces operational efficiency.


To address the frequency mismatch between high-level reasoning and physical control, hierarchical frameworks have been introduced \cite{airhunt, vla-an, dualvln, hirobot}. For instance, AirHunt \cite{airhunt} decouples the construction of a semantic 3D value map from path planning to mitigate the stop-and-infer issue. While integrating VLMs to guide continuous object search, this method inherently relies on a rigid, problem-specific pipeline and fails to accommodate aperiodic user interventions during physical execution. To address dynamic commands, Hi-Robot \cite{hirobot} offers a hierarchical alternative by enabling a high-level VLM to asynchronously update natural language instructions for a low-level VLA, achieving seamless instruction integration. However, this approach necessitates a robust VLA model and the fine-tuning of the upper-level VLM. 

Recently, several approaches have explored spatial reasoning using 3D scene graphs. For example, ConceptGraphs \cite{gu2024conceptgraphs} construct object-centric 3D graphs using open-vocabulary segmentation and dense 3D mapping, and leverage LLMs to infer spatial relationships. Beyond object-level reasoning, subsequent works organize objects into manually defined hierarchical structures \cite{yin2024sg, yin2025unigoal, zhang2026spatialnav}.
However, constructing such 3D scene graphs typically requires dense reconstruction of the environment for subsequent navigation, which introduces substantial computational overhead. Moreover, when spatial reasoning involves objects not predefined in the graph vocabulary, the graph structure may need extensive updates to accommodate new semantic categories.



Unlike existing training-free approaches, we present a responsive agent framework that decouples high-level reasoning from physical execution. The key to our system is its asynchronous multi-agent architecture: Foreground Workflow Agents handle active tasks and user interactions, while Background Agents perform look-ahead reasoning for upcoming tasks. The framework relies on a lightweight topological map built from keyframe images, avoiding dense 3D reconstruction. This concurrent design enables complex semantic reasoning and hierarchical task planning while the quadrotor executes agile flight maneuvers, effectively bridging high-level cognition with low-level control.

\section{Methodology}
\subsection{System Overview}

To enable long-horizon semantic navigation with non-blocking interaction in partially known environments, we develop the agent system as shown in Fig. \ref{fig:system_overview}. The system adopts a two-layer architecture that separates cognitive reasoning from physical execution. The cognitive layer runs on a ground station, while the physical layer operates onboard the UAV.
We first elaborate on the cognitive layer (Sec. \ref{sec:cognitive}), which generates high-level decisions and provides commands to the physical layer. Next, we introduce the Impression Graph, a topological representation for reasoning and navigation (Sec. \ref{sec:impression_graph}). Finally, we present the physical layer, which performs collision avoidance and trajectory tracking (Sec. \ref{sec:physical}).

\begin{figure}
  \centering
  \includegraphics[width=0.49\textwidth]{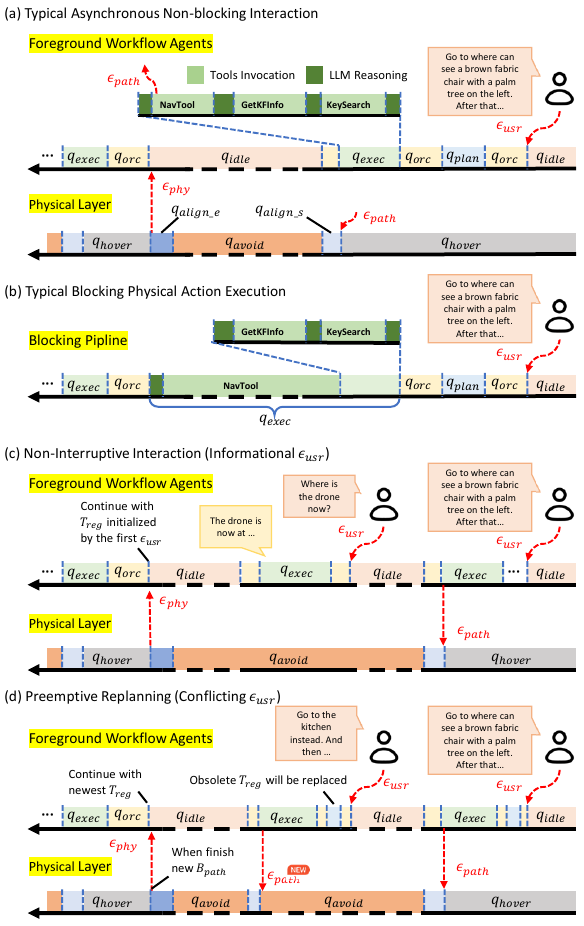}
  \caption{\textbf{Timelines of Typical Cases.} 
In each sub-image, the upper rows depict the Foreground Workflow Agents' lifecycle, where $q_{idle}$ marks the idle state, $q_{orc}$ indicates the orchestrator's active routing phase, and $q_{plan}/q_{exec}$ denote the engagement of the planner and executor, respectively. The lower rows track the physical layer states.
\textbf{(a) Our suspend-and-resume protocol} yields the agent to the idle state ($q_{idle}$) immediately after triggering $\epsilon_{path}$, enabling ``Chatting-while-Flying'', unlike the \textbf{Blocking Baseline (b)} where the agent is blocked until physical completion.
\textbf{(c) Informational Queries} are resolved in parallel with the ongoing flight state ($q_{avoid}$).
\textbf{(d) Conflicting Commands} trigger preemptive replanning, issuing a new $\epsilon_{path}$ to immediately redirect the UAV.}
  \label{fig:timeline}
\end{figure}

\subsection{Cognitive Layer}
\label{sec:cognitive}
The cognitive layer consists of three components: Foreground Workflow Agents, Background Agents, and a skill library. The Foreground Workflow Agents manage the active interaction loop and handle user-issued natural language commands that may arrive at arbitrary times. The Background Agents pre-execute upcoming sub-tasks within a mission queue, which stores the ordered sub-tasks of the current mission. This design reduces idle hovering caused by reasoning delays and enables non-blocking interaction. Both types of agents invoke the skill library to execute task-specific operations.

\subsubsection{Foreground Workflow Agents}

The Foreground Workflow Agents implement three roles: an \emph{orchestrator} that routes control flow, a \emph{planner} that decomposes high-level instructions into conditional task trees, and an \emph{executor} implemented as a scoped ReAct agent \cite{yao2022react} for atomic sub-tasks. Unlike serial architectures with blocking tool-use logic, this workflow follows an event-driven suspend-and-resume protocol.

When the \emph{executor} invokes a physical actuation skill, it sends the planned actions or path to the onboard physical layer and immediately yields control. As shown in Fig. \ref{fig:timeline}a, the Foreground Workflow Agents then enter an idle state while the physical layer handles the actual flight. This suspension allows the \emph{orchestrator} to monitor user commands $\epsilon_{usr}$ and physical feedback $\epsilon_{phy}$ in parallel with the ongoing motion.

The system handles responses dynamically to the semantic intent of the user command $\epsilon_{usr}$. For informational queries, such as requesting the current pose of the UAV as shown in Fig. \ref{fig:timeline}c, the \emph{orchestrator} invokes perceptual skills in parallel with flight execution, maintaining the trajectory uninterrupted. For conflicting commands that specify a new goal, the system triggers a preemptive replanning mechanism as shown in Fig. \ref{fig:timeline}d. The \emph{orchestrator} directs the planner to discard the obsolete task registry and generate a new sequence, effectively overwriting the physical trajectory. This design allows the agent to handle both context queries and goal modifications during flight.

\begin{figure}
  \centering
  \includegraphics[width=0.49\textwidth]{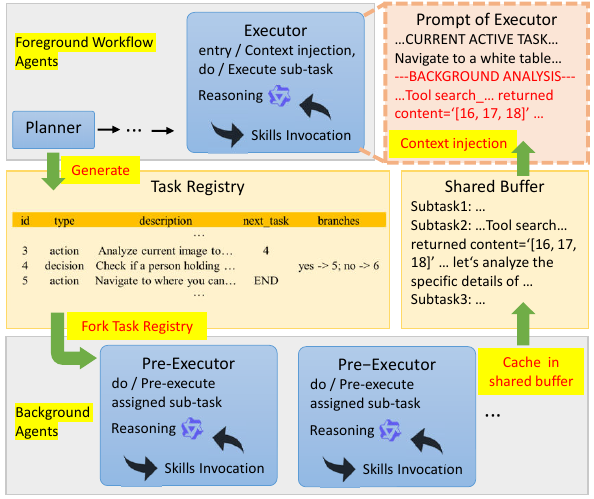}
  \caption{\textbf{Data Flow of Background Pre-execution.} 
  The pipeline branches from the task registry (center left). While the Foreground Workflow Agents execute the task from the first one, upcoming sub-tasks are assigned across Background Agents (bottom) for pre-execution using mnemonic skills. The retrieved context is cached in a shared buffer (center right) and dynamically injected into the prompt of the Executor (top right).}
  \label{fig:background}
\end{figure}

\subsubsection{Background Agents}

To reduce the inference latency inherent in sequential reasoning, the system employs a background look-ahead mechanism as shown in Fig. \ref{fig:background}.
After the \emph{planner} generates a task sequence, the system asynchronously dispatches the execution to Background Agents. While the Foreground Workflow Agents process the current sub-task, the Background Agents analyze upcoming sub-tasks in the mission queue.

These Background Agents act as predictive workers limited to information retrieval. To ensure safety and logical consistency, their actions are confined to querying the Impression Graph using mnemonic skills. 

The results of this pre-execution are stored in a shared buffer. When the Foreground Workflow Agents advance to a pre-analyzed sub-task, the system retrieves this cached context and injects it into the executor prompt. This context injection mechanism masks the latency of scene prior retrieval, reducing the hovering time and enabling smoother transitions between navigation and decision-making phases compared with stop-and-infer systems.

\subsubsection{Skill Library}

The skill library contains modules invoked on demand by the agents, providing mnemonic retrieval, real-time perception, and navigation capabilities.
By separating reasoning from specific implementations, the cognitive layer focuses on high-level decision-making, while specialized modules handle low-level execution.

\emph{Mnemonic skills} enable agents to interact with the Impression Graph (Sec. ~\ref{sec:impression_graph}) and access scene priors efficiently. Depending on the specific function invoked, the system performs semantic matching, information retrieval, or visual reasoning on the graph nodes. Unlike approaches \cite{xiao2025uav, kathirvel2025sent} that inject massive environmental context into the language model prompt via a fixed format, our architecture enforces an on-demand retrieval paradigm. This design decouples environmental complexity from the context length limits of the language model. Furthermore, it allows specific simple functions, such as fuzzy retrieval or visual captioning, to be implemented by lightweight models. This optimizes inference speed and deployment costs while allowing queries only when needed.

\emph{Perceptual skills} provide access to real-time information. When invoked, the agent can read the UAV's current state or request a vision-language model (VLM) to analyze the current RGB camera image. This ensures that reasoning remains grounded in the immediate physical environment.

The \emph{navigation skill} executes physical transit to keyframe nodes.
When invoked, the system maps the UAV’s current position to the nearest node on the Impression Graph and computes the shortest topological path to the target keyframe.
This topological sequence is projected into world coordinates to generate a series of 3D waypoints with a final orientation.
The resulting trajectory event serves as a non-blocking trigger for the physical layer, allowing the cognitive workflow to suspend immediately and await physical feedback or new user commands.

\subsection{Impression Graph}
\label{sec:impression_graph}

We construct the Impression Graph $\mathcal{G}_{imp} = (\mathcal{N}, \mathcal{E})$ as a sparse topological representation of scene priors. The node set $\mathcal{N}$ consists of keyframes, where each node $n_i \in \mathcal{N}$ contains a multimodal context tuple including visual streams, world pose, CLIP \cite{clip} embeddings, and generated semantic captions. The edge set $\mathcal{E}$ encodes navigability: a connection between nodes indicates traversability for the UAV and is weighted by the physical Euclidean distance to support path planning. The incremental construction process transforms a continuous posed-RGBD stream into this topological map through three stages: \emph{node creation}, \emph{connectivity establishment}, and \emph{semantic enrichment}.

\emph{Node creation}: To maintain graph compactness, we employ a spatio-visual selection strategy. The system registers a new frame as a node $n_i$ by comparing it against existing nodes within a predefined search radius. A new node is created only if it shows significant spatial displacement (i.e., Euclidean distance or yaw difference exceeding thresholds) or visual novelty (via CLIP cosine similarity) relative to these spatially proximal frames.

\begin{figure}
  \centering
  \includegraphics[width=0.49\textwidth]{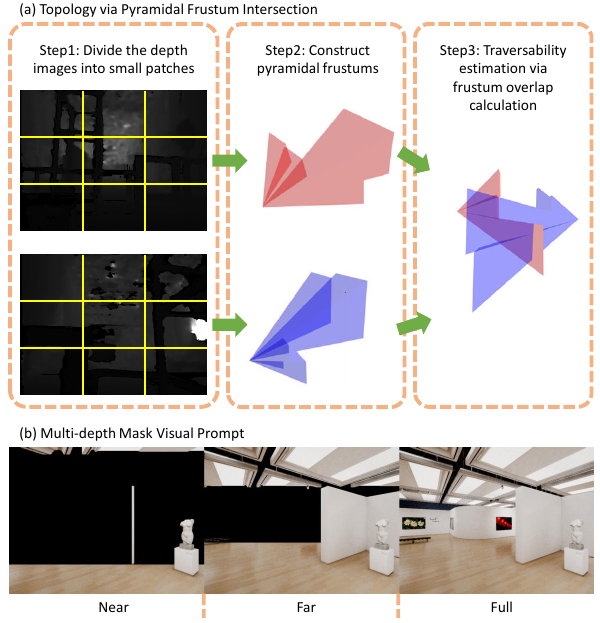}
  \caption{\textbf{Impression Graph Construction Pipeline.} 
  (a) \textbf{Topological Connectivity:} The depth map is tessellated into patches and projected into geometric pyramidal frustums. Edges $(n_i, n_j)$ are established solely if the volumetric intersection of their frustums exceeds $\sigma_{vol}$.
  (b) \textbf{Semantic Generation:} $I_{rgb}$ is segmented into depth-stratified views (Near, Far, Full) and concatenated into a composite input for the VLM.}
  \label{fig:impression_graph}
\end{figure}

\emph{Connectivity establishment}: Edges in $\mathcal{E}$ encode navigability. To reduce the safety risks of simple distance-based heuristics and the computational latency of dense reconstruction, we adopt a lightweight pyramidal frustum intersection criterion as shown in Fig. \ref{fig:impression_graph}a. The depth map is tessellated into patches, each projected into a geometric pyramid defined by its local minimum depth, approximating the view frustum as a set of pyramids. An edge is created between two nodes only if the volumetric intersection of their pyramid sets exceeds a predefined threshold $\sigma_{vol}$. Unlike methods that infer connectivity from data collection paths \cite{kathirvel2025sent}, our method can discover novel connections, such as aerial shortcuts over ground-level barriers, even if those paths were not traversed during the initial data collection.

\emph{Semantic enrichment}: Each node is enhanced with a structured semantic description using a multi-depth masked prompting pipeline as shown in Fig. \ref{fig:impression_graph}b. Instead of processing the raw image directly, the system segments the RGB input into near, far, and full views based on depth thresholds. These masked views are concatenated into a composite input, adopting a visual stitching strategy \cite{xu2024vlm-grounder}. A VLM \cite{bai2025qwen3}  processes this composite to generate spatially aware captions. The multi-depth approach reduces hallucinations and enables accurate distinction between foreground objects and background landmarks.

\subsection{Physical Layer}
\label{sec:physical}
The physical layer executes the generated trajectory through a robust onboard motion controller. This controller manages the finite state machine (as shown in Fig. \ref{fig:system_overview}) and tracks the given path $\epsilon_{path}$ via a collision avoidance policy. To ensure safety during high-speed flight, we employ a neural reactive policy $\pi_\theta$ parameterized by a convolutional recurrent neural network and trained via differentiable physics \cite{zhang2025learning}:
\begin{equation}
    \mathbf{a}_t, \mathbf{h}_t = \pi_\theta(I_{depth}, \mathbf{s}_t, \mathbf{h}_{t-1}).
\end{equation}
Operating at 30 Hz, the policy maps the depth stream $I_{depth}$, vehicle state $\mathbf{s}_t$, and a temporal memory $\mathbf{h}_{t-1}$ to control actions $\mathbf{a}_t$ at each timestep $t$.

\section{Experiments}
\begin{table*}
\centering
\begin{threeparttable}
\caption{\textbf{Quantitative Analysis in Simulation.}
$\uparrow$: higher is better, $\downarrow$: lower is better. 
The best results are highlighted in \textbf{bold}.}
\label{tab:sim_results}
\small 
\begin{tabular}{l|cccc|cc|c}
\toprule
\textbf{Method} & \textbf{SR} (\%) $\uparrow$ & \textbf{OSR} (\%) $\uparrow$ & \textbf{Prog.} (\%) $\uparrow$ & \textbf{SPL} $\uparrow$ & \textbf{TET} (s) $\downarrow$ & \textbf{HT} (s) $\downarrow$ & \textbf{IA} (\%) $\uparrow$ \\ 
\midrule
Serial Plan-ReAct \cite{uav-codeagent} & 78.3 & 83.3 & 92.0 & 57.2 & 470.2 & 249.9 & 0.0 \\
Scene Graph Agent \cite{gu2024conceptgraphs} & 15.0 & 18.3 & 40.7 & 11.9 & 338.6 & 170.8 & 24.9 \\
Traj NavGraph \cite{kathirvel2025sent} & 76.7 & 85.0 & 92.2 & 49.8 & 379.2 & 125.8 & 57.5 \\ 
\midrule
w/o Look-ahead & 76.7 & 81.7 & 90.3 & 55.5 & 445.2 & 226.0 & 38.2 \\
w/o On-demand Retrieval & 60.0 & 68.3 & 78.8 & 42.5 & \textbf{335.0} & \textbf{113.7} & 52.6 \\ 
\midrule
\textbf{QuadAgent (Ours)} & \textbf{80.0} & \textbf{86.7} & \textbf{93.3} & \textbf{58.6} & 353.9 & 118.0 & \textbf{58.3} \\ 
\bottomrule
\end{tabular}

\end{threeparttable}
\end{table*}

Our experiments address four core research questions:

\textbf{RQ1:} How effectively does the system maintain non-blocking responsiveness during physical execution and maximize the interactive time window?

\textbf{RQ2:} Can the asynchronous look-ahead architecture mask inference latency to minimize invalid hovering time?

\textbf{RQ3:} Do mnemonic skills effectively retrieve from the Impression Graph to ground natural language requests into physical targets?

\textbf{RQ4:} Does the topology of the Impression Graph, built via frustum intersection, minimize trajectory lengths and enable the discovery of aerial shortcuts?

To address these questions, we conduct quantitative benchmarks in simulations (Sec. \ref{sec:exp_sim}) and qualitative case studies on a real-world platform (Sec. \ref{sec:exp_real}).

\subsection{Quantitative Analysis in Simulation}
\label{sec:exp_sim}

\noindent \textbf{Environments and Implementations.} Existing navigation benchmarks predominantly rely on discrete action spaces or timesteps. They structurally ignore continuous UAV kinematics, lack metrics to measure the interactive time window, and rarely feature long-horizon missions governed by conditional logic. To bridge these gaps, we conduct experiments in three challenging environments within the AirSim simulator: an art gallery, a downtown street, and a big office. The evaluation dataset comprises 60 long-horizon tasks, with $25\%$ requiring conditional logical reasoning. Regarding implementation, our framework is powered by the Qwen model suite via API (Qwen-Max for primary reasoning, Qwen-Plus/Qwen-VL-Plus for mnemonic and perceptual skills, and Qwen-VL-Max for node semantics).

\noindent \textbf{Baselines.} To validate the performance of QuadAgent, we compare our approach against three representative baselines derived from prevailing technical paradigms. First, \textit{Serial Plan-ReAct} instantiates the serial tool-use paradigm prevalent in current aerial LLM agents (e.g., UAV Agent in \cite{uav-codeagent}), invoking navigation skills in a blocking manner. To ensure a rigorous and fair comparison, we specifically tailored this paradigm to our task formulation by equipping it with the identical skill library. Second, \textit{Scene Graph Agent} represents the scene graph reasoning paradigm. This approach constructs scene priors using ConceptGraphs \cite{gu2024conceptgraphs}, which serves as the foundational representation in recent state-of-the-art indoor navigation methods \cite{yin2024sg, yin2025unigoal}. To ensure a fair comparison, we utilize Qwen-VL-Max for its node captions—identical to our framework—and augment its nodes with ground-truth textual identifiers (e.g., storefront names). Since pure scene graphs lack physical navigability, the agent navigates to the reasoned targets using the same navigation method employed in our framework. Third, \textit{Traj NavGraph} represents a trajectory-based topological navigation scheme. Node connectivity follows a hybrid rule set where edges are primarily established based on the movement trajectory of the operator during data collection, similar to SENT-Map \cite{kathirvel2025sent}; to enhance local connectivity, we add a connecting edge when the Euclidean distance between two nodes falls below a proximity threshold.

\noindent \textbf{Ablation Studies.} To verify the effectiveness of our architectural components, we evaluate two variants. The \textit{w/o Look-ahead} variant disables the background pre-executor, forcing the agent to operate in a stop-and-infer mode. The \textit{w/o On-demand Retrieval} variant removes the dynamic tool invocation retrieval mechanism, replacing it with a fixed context window. In this setup, we directly inject the 20 nodes most relevant to the task CLIP vector, along with the real-time state and image, into the prompt for each sub-task.

\noindent \textbf{Metrics.} Building upon previous work \cite{an2024etpnav}, we adapt standard evaluation metrics to accommodate sequential tasks. Specifically, we use Success Rate (SR) (the percentage of episodes where the agent visits all sequential targets in the strict correct order and properly terminates) and Oracle Success Rate (OSR) (whether the agent physically passes through all target zones during the flight, regardless of explicit stop commands). To assess routing efficiency, we use Success weighted by Path Length (SPL), uniquely defining the optimal path as the cumulative Euclidean distance between sequential targets to strictly penalize detouring behaviors. 
Additionally, to capture the degree of completion in long-horizon missions, we introduce Progress (Prog.), defined as the average fraction of sequential targets successfully reached in the correct order per episode. We report Total Execution Time (TET) and Hovering Time (HT) to assess system efficiency. Finally, we measure non-blocking interactivity using Interaction Availability (IA), defined as the ratio of the duration during which the system remains responsive to user inputs to the total task time.

\begin{figure*}
\centering
\includegraphics[width=0.99\textwidth]{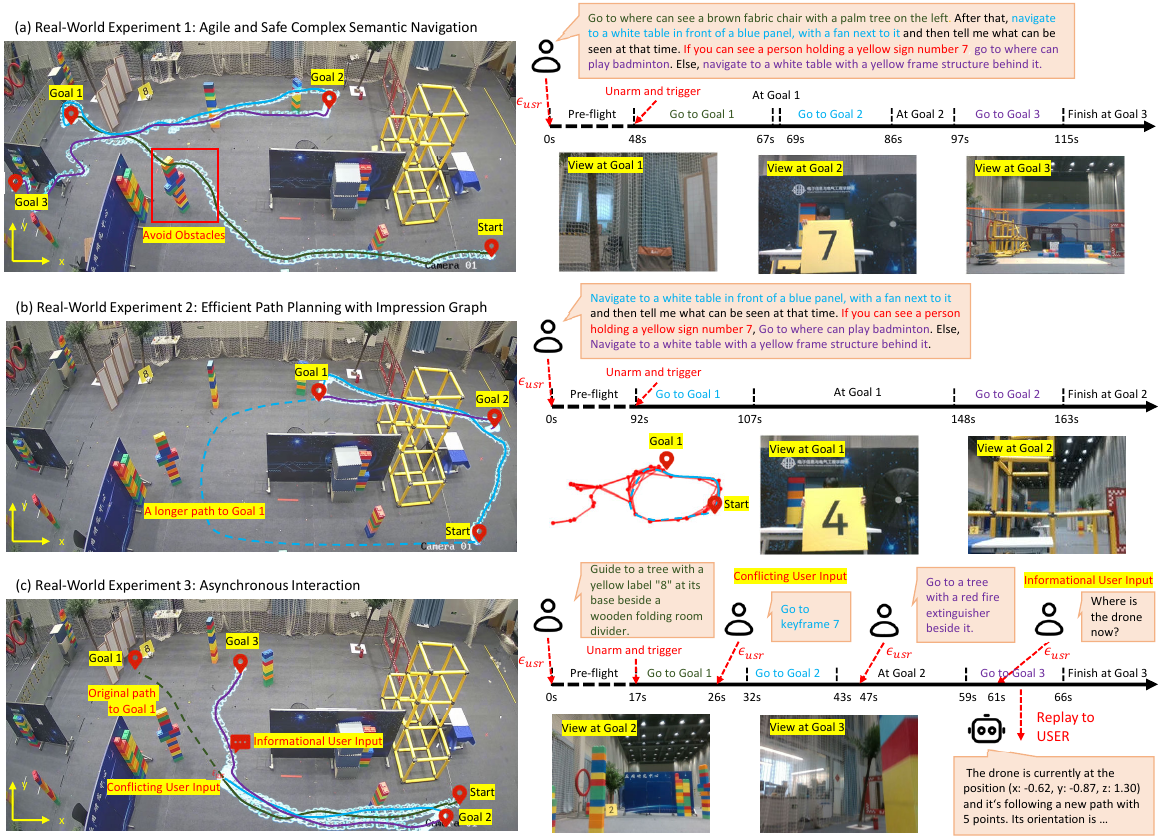}
\caption{\textbf{Qualitative evaluation of the QuadAgent in real-world environments.} \textbf{(a) Experiment 1:} The UAV executes a long-horizon conditional task. The trajectory highlights an agile avoidance of unmapped obstacles (red box) and the timeline illustrates the autonomous sequential completion of sub-tasks. \textbf{(b) Experiment 2:} The agent utilizes the topological connectivity of the Impression Graph to identify and execute a spatial shortcut (solid blue line) to Goal 1 compared with the longer one (dashed blue line). \textbf{(c) Experiment 3:} During the navigation to Goal 1, a conflicting user input triggers an immediate trajectory update towards Goal 2 (blue line), preempting the original planned path to Goal 1 (dark green dashed line). While navigating to Goal 3, the agent processes an informational user input, answering the user in parallel without interrupting the physical flight. Note: Text color-coding and goal annotations are for visualization only. In practice, the agent receives a single unified instruction and infers targets autonomously.}
\label{fig:real-world-exp}
\end{figure*}

QuadAgent successfully addresses RQ1 and RQ2 by dramatically improving interaction availability and reducing hovering time. Compared to the completely blocking \textit{Serial Plan-ReAct} (0.0\% IA, 249.9s HT), our architecture achieves 58.3\% IA and halves the HT to 118.0s. Furthermore, \textit{w/o Look-ahead} nearly doubles the HT to 226.0s and degrades IA to 38.2\%. This proves that the asynchronous look-ahead mechanism is indispensable for masking latency and improving non-blocking interaction.

Answering RQ4, while the \textit{Traj NavGraph} baseline achieves a competitive OSR, its SPL drops to 49.8. By discovering shortcuts with frustum intersection, QuadAgent elevates the SPL to 58.6, showing superior path planning.

Regarding RQ3, QuadAgent achieves the highest SR of 80.0\% and OSR of 86.7\%. The \textit{Scene Graph Agent} baseline, despite being augmented with ground-truth textual identifiers and the same Qwen backbone, exhibits a substantial overall performance decline (15.0\% SR and 40.7\% Prog.). Notably, while achieving 45.0\% SR and 61.0\% Prog. in the structured art gallery, this baseline fails completely in downtown and office scenes. This reveals that predefined scene graphs struggle in unbounded outdoor spaces and environments with highly repetitive objects. Furthermore, \textit{w/o On-demand Retrieval} causes a precipitous SR drop to 60.0\% and Prog. to 78.8\%; although bypassing the multi-turn dialogue yields the lowest TET (335.0s), this extreme trade-off severely compromises fundamental task completion capabilities.

\subsection{Real-World Evaluation}
\label{sec:exp_real}


We deployed QuadAgent on a custom-built fpv-style UAV equipped with a resource-constrained Radxa Zero 3W SBC and an Intel RealSense D435i (Fig. \ref{fig:system_overview}). The 10m$\times$25m arena features diverse obstacles and unmapped dynamic blocks to test physical robustness. The sparse Impression Graph was derived from a handheld RGB-D stream.

\noindent \textbf{Agile Navigation \& Reasoning (RQ2 \& RQ3).} 
To validate performance under physical constraints, the agent executed a sequential conditional mission (Fig. \ref{fig:real-world-exp}a). Throughout the missions, the differentiable policy $\pi_\theta$ ensured robust avoidance of unmapped obstacles at peak speeds of 5.10 m/s, confirming physical agility.
Leveraging background look-ahead, the system pre-localized subsequent targets during the flight phase. Through Context Injection, this parallel reasoning reduced inference-induced hovering to merely 2 seconds at Goal 1 in Experiment 1.
Upon visual confirmation of the condition (``a person holding a yellow sign number 7''), the agent autonomously routed to the correct branch. 

\noindent \textbf{Discovery of Shortcuts (RQ4).} Notably, the Impression Graph enabled the discovery of a shortcut over a scaffolding frame—a path validated by frustum intersection but inaccessible to human data collectors (Experiment 2 in Fig. \ref{fig:real-world-exp}b).

\noindent \textbf{Asynchronous Interaction (RQ1).} 
We further validated non-blocking interaction via two dynamic interrupts in Experiment 3 (Fig. \ref{fig:real-world-exp}c). 
First, a conflicting user input was issued during transit to Goal 1. The agents immediately updated the task registry, seamlessly redirecting the UAV to Goal 2. 
Second, an informational user input was processed during navigation to Goal 3. The system generated natural language feedback while the physical layer continued trajectory tracking without interruption.



\section{Conclusion and Discussion}
This paper presents QuadAgent, an asynchronous multi-agent architecture that overcomes the latency and stop-and-infer limitations of serial LLM-based agents. By enabling asynchronous execution with background look-ahead reasoning and suspend-and-resume protocols, the system achieves responsive, non-blocking interaction during agile flight. The proposed Impression Graph provides lightweight scene priors that replace hesitant exploration with structured and confident navigation. With these priors, the UAV reaches real-world flight speeds of up to 5 m/s, significantly surpassing prior approaches that are typically limited to below 2 m/s. A vision-based physical safety layer further ensures robust obstacle avoidance, even under network instability or LLM failures. In future work, we will expand the agent’s skill library to support autonomous online exploration, richer physical interactions, and more general active perception capabilities.









\bibliographystyle{IEEEtran}
\bibliography{refs}

@inproceedings{gu2024conceptgraphs,
  title={Conceptgraphs: Open-vocabulary 3d scene graphs for perception and planning},
  author={Gu, Qiao and Kuwajerwala, Ali and Morin, Sacha and others},
  booktitle={2024 IEEE International Conference on Robotics and Automation (ICRA)},
  pages={5021--5028},
  year={2024},
  organization={IEEE}
}

@article{yin2024sg,
  title={Sg-nav: Online 3d scene graph prompting for llm-based zero-shot object navigation},
  author={Yin, Hang and Xu, Xiuwei and Wu, Zhenyu and others},
  journal={Advances in Neural Information Processing Systems},
  volume={37},
  pages={5285--5307},
  year={2024}
}

@misc{wang2025uavflowcolosseorealworldbenchmark,
      title={UAV-Flow Colosseo: A Real-World Benchmark for Flying-on-a-Word UAV Imitation Learning}, 
      author={Xiangyu Wang and Donglin Yang and Yue Liao and others},
      year={2025},
      eprint={2505.15725},
      archivePrefix={arXiv},
      primaryClass={cs.RO},
      url={https://arxiv.org/abs/2505.15725}, 
}

@inproceedings{hu2025see,
  title={See, Point, Fly: A Learning-Free VLM Framework for Universal Unmanned Aerial Navigation},
  author={Hu, Chih Yao and Lin, Yang-Sen and Lee, Yuna and others},
  booktitle={Conference on Robot Learning},
  pages={4697--4708},
  year={2025},
  organization={PMLR}
}

@article{li2025skyvln,
  title={SkyVLN: Vision-and-Language Navigation and NMPC Control for UAVs in Urban Environments},
  author={Li, Tianshun and Huai, Tianyi and Li, Zhen and others},
  journal={arXiv preprint arXiv:2507.06564},
  year={2025}
}

@inproceedings{xiao2025uav,
  title={Uav-on: A benchmark for open-world object goal navigation with aerial agents},
  author={Xiao, Jianqiang and Sun, Yuexuan and Shao, Yixin and others},
  booktitle={Proceedings of the 33rd ACM International Conference on Multimedia},
  pages={13023--13029},
  year={2025}
}

@inproceedings{wu2025selp,
  title={SELP: Generating safe and efficient task plans for robot agents with large language models},
  author={Wu, Yi and Xiong, Zikang and Hu, Yiran and others},
  booktitle={2025 IEEE International Conference on Robotics and Automation (ICRA)},
  pages={2599--2605},
  year={2025},
  organization={IEEE}
}

@article{vla-an,
  title={VLA-AN: An Efficient and Onboard Vision-Language-Action Framework for Aerial Navigation in Complex Environments},
  author={Wu, Yuze and Zhu, Mo and Li, Xingxing and others},
  journal={arXiv preprint arXiv:2512.15258},
  year={2025}
}

@article{zhang2025learning,
  title={Learning vision-based agile flight via differentiable physics},
  author={Zhang, Yuang and Hu, Yu and Song, Yunlong and others},
  journal={Nature Machine Intelligence},
  pages={1--13},
  year={2025},
  publisher={Nature Publishing Group UK London}
}

@article{xu2024vlm-grounder,
  title={Vlm-grounder: A vlm agent for zero-shot 3d visual grounding},
  author={Xu, Runsen and Huang, Zhiwei and Wang, Tai and others},
  journal={arXiv preprint arXiv:2410.13860},
  year={2024}
}

@inproceedings{yao2022react,
  title={React: Synergizing reasoning and acting in language models},
  author={Yao, Shunyu and Zhao, Jeffrey and Yu, Dian and others},
  booktitle={The eleventh international conference on learning representations},
  year={2022}
}

@misc{pi0.5,
      title={$\pi_{0.5}$: a Vision-Language-Action Model with Open-World Generalization}, 
      author={Physical Intelligence and Kevin Black and Noah Brown and others},
      year={2025},
      eprint={2504.16054},
      archivePrefix={arXiv},
      primaryClass={cs.LG},
      url={https://arxiv.org/abs/2504.16054}, 
}

@misc{dualvln,
      title={Ground Slow, Move Fast: A Dual-System Foundation Model for Generalizable Vision-and-Language Navigation},
      author={Meng Wei and Chenyang Wan and Jiaqi Peng and others},
      year={2025},
      eprint={2512.08186},
      archivePrefix={arXiv},
      primaryClass={cs.RO},
      url={https://arxiv.org/abs/2512.08186},
}

@article{navfom,
  title={Embodied navigation foundation model},
  author={Zhang, Jiazhao and Li, Anqi and Qi, Yunpeng and others},
  journal={arXiv preprint arXiv:2509.12129},
  year={2025}
}

@article{OpenFly,
  author       = {Yunpeng Gao and
                  Chenhui Li and
                  Zhongrui You and
                  and others},
  title        = {OpenFly: A Comprehensive Platform for Aerial Vision-Language Navigation},
  journal      = {CoRR},
  volume       = {abs/2502.18041},
  year         = {2025}
}

@misc{uss-nav,
      title={USS-Nav: Unified Spatio-Semantic Scene Graph for Lightweight UAV Zero-Shot Object Navigation}, 
      author={Weiqi Gai and Yuman Gao and Yuan Zhou and others},
      year={2026},
      eprint={2602.00708},
      archivePrefix={arXiv},
      primaryClass={cs.RO},
      url={https://arxiv.org/abs/2602.00708}, 
}

@misc{airhunt,
      title={AirHunt: Bridging VLM Semantics and Continuous Planning for Efficient Aerial Object Navigation}, 
      author={Xuecheng Chen and Zongzhuo Liu and Jianfa Ma and others},
      year={2026},
      eprint={2601.12742},
      archivePrefix={arXiv},
      primaryClass={cs.RO},
      url={https://arxiv.org/abs/2601.12742}, 
}

@inproceedings{pmlr-v205-ichter23a,
  title={Do as i can, not as i say: Grounding language in robotic affordances},
  author={Brohan, Anthony and Chebotar, Yevgen and Finn, Chelsea and others},
  booktitle={Conference on robot learning},
  pages={287--318},
  year={2023},
  organization={PMLR}
}

@misc{roboos,
      title={RoboOS: A Hierarchical Embodied Framework for Cross-Embodiment and Multi-Agent Collaboration}, 
      author={Huajie Tan and Xiaoshuai Hao and Cheng Chi and others},
      year={2025},
      eprint={2505.03673},
      archivePrefix={arXiv},
      primaryClass={cs.RO},
      url={https://arxiv.org/abs/2505.03673}, 
}

@misc{uav-codeagent,
      title={UAV-CodeAgents: Scalable UAV Mission Planning via Multi-Agent ReAct and Vision-Language Reasoning}, 
      author={Oleg Sautenkov and Yasheerah Yaqoot and Muhammad Ahsan Mustafa and others},
      year={2025},
      eprint={2505.07236},
      archivePrefix={arXiv},
      primaryClass={cs.RO},
      url={https://arxiv.org/abs/2505.07236}, 
}

@misc{hirobot,
      title={Hi Robot: Open-Ended Instruction Following with Hierarchical Vision-Language-Action Models}, 
      author={Lucy Xiaoyang Shi and Brian Ichter and Michael Equi and others},
      year={2025},
      eprint={2502.19417},
      archivePrefix={arXiv},
      primaryClass={cs.RO},
      url={https://arxiv.org/abs/2502.19417}, 
}

@inproceedings{yin2025unigoal,
  title={Unigoal: Towards universal zero-shot goal-oriented navigation},
  author={Yin, Hang and Xu, Xiuwei and Zhao, Linqing and others},
  booktitle={Proceedings of the Computer Vision and Pattern Recognition Conference},
  pages={19057--19066},
  year={2025}
}

@article{kathirvel2025sent,
  title={SENT Map--Semantically Enhanced Topological Maps with Foundation Models},
  author={Kathirvel, Raj Surya Rajendran and Chavis, Zach A and Guy, Stephen J and Desingh, Karthik},
  journal={arXiv preprint arXiv:2511.03165},
  year={2025}
}

@article{an2024etpnav,
  title={Etpnav: Evolving topological planning for vision-language navigation in continuous environments},
  author={An, Dong and Wang, Hanqing and Wang, Wenguan and others},
  journal={IEEE Transactions on Pattern Analysis and Machine Intelligence},
  year={2024},
  publisher={IEEE}
}

@article{zhang2026spatialnav,
  title={SpatialNav: Leveraging Spatial Scene Graphs for Zero-Shot Vision-and-Language Navigation},
  author={Zhang, Jiwen and Li, Zejun and Wang, Siyuan and others},
  journal={arXiv preprint arXiv:2601.06806},
  year={2026}
}

@article{kim2024openvla,
  title={Openvla: An open-source vision-language-action model},
  author={Kim, Moo Jin and Pertsch, Karl and Karamcheti, Siddharth and Xiao, Ted and Balakrishna, Ashwin and Nair, Suraj and Rafailov, Rafael and Foster, Ethan and Lam, Grace and Sanketi, Pannag and others},
  journal={arXiv preprint arXiv:2406.09246},
  year={2024}
}

@article{pi0,
  title={{$\pi_0$}: A Vision-Language-Action Flow Model for General Robot Control},
  author={Black, Kevin and Brown, Noah and Driess, Danny and others},
  journal={arXiv preprint arXiv:2410.24164},
  year={2024}
}

@article{manus,
  title={From mind to machine: The rise of manus ai as a fully autonomous digital agent},
  author={Shen, Minjie and Li, Yanshu and Chen, Lulu and Yang, Qikai},
  journal={arXiv preprint arXiv:2505.02024},
  year={2025}
}

@misc{openclaw,
  author = {Steinberger, Peter and others},
  title = {OpenClaw — Personal AI Assistant},
  year = {2025},
  publisher = {GitHub},
  journal = {GitHub repository},
  howpublished = {\url{https://github.com/openclaw/openclaw}}
}

@article{bai2025qwen3,
  title={Qwen3-vl technical report},
  author={Bai, Shuai and Cai, Yuxuan and Chen, Ruizhe and others},
  journal={arXiv preprint arXiv:2511.21631},
  year={2025}
}

@inproceedings{clip,
  title={Learning transferable visual models from natural language supervision},
  author={Radford, Alec and Kim, Jong Wook and Hallacy, Chris and others},
  booktitle={International conference on machine learning},
  pages={8748--8763},
  year={2021},
  organization={PmLR}
}

@article{pardyl2025flysearch,
  title={FlySearch: Exploring how vision-language models explore},
  author={Pardyl, Adam and Matuszek, Dominik and Przebieracz, Mateusz and Cygan, Marek and others},
  journal={arXiv preprint arXiv:2506.02896},
  year={2025}
}

\end{document}